# A Proposal for Extending the Common Model of Cognition to Emotion

**Paul S. Rosenbloom[1] (Rosenbloom@USC.edu), John E. Laird[2] (John.Laird@cic.iqmri.org), Christian Lebiere[3] (CL@CMU.edu), Andrea Stocco[4] (Stocco@UW.edu), Richard H. Granger[5] (Richard.Granger@Dartmouth.edu) & Christian Huyck[6] (C.Huyck@MDX.ac.uk)**

[1] Institute for Creative Technologies & Thomas Lord Dept. of Computer Science, University of Southern California
[2] Center for Integrated Cognition, IQMRI
[3] Department of Psychology, Carnegie Mellon University
[4] Department of Psychology, University of Washington
[5] Department of Psychological and Brain Sciences, Dartmouth University
[6] Department of Computer Science, Middlesex University

**Abstract**

Cognition and emotion must be partnered in any complete model of a humanlike mind. This article proposes an extension to the Common Model of Cognition – a developing consensus concerning what is required in such a mind – for emotion that includes a linked pair of modules for emotion and metacognitive assessment, plus pervasive connections between these two new modules and the Common Model's existing modules and links.

## Introduction

The *Common Model of Cognition* (Rosenbloom, Lebiere & Laird, 2022) – née the *Standard Model of the Mind* (Laird, Lebiere & Rosenbloom, 2017) – is a developing consensus concerning what must be in a cognitive architecture to support *humanlike minds*. The consensus is derived from existing cognitive architectures, from researchers who study them, and from results relevant to them, with humanlike minds comprising human minds plus any other natural or artificial minds similar enough to be modellable in the same manner at the chosen level of abstraction.

The Common Model is not intended to be a cognitive architecture in the traditional sense, in being abstract, radically incomplete, and not directly executable. What it includes is limited to what the community can reach a consensus on concerning its necessity for humanlike cognition. Sufficiency considerations play into what topics are considered for consensus building but play no direct role in judging what is actually to be included.

This article reports on an effort to address one major source of incompleteness in the Common Model – concerning *emotion* – that has not yet reached a consensus with respect to necessity. It thus amounts to a proposal for how to extend the Common Model to particular aspects of emotion but not (yet) an actual extension of the Common Model to emotion.

As stated in Larue et al. (2018), "Modeling emotion is essential to the Common Model of Cognition … because emotion can't be divorced from cognition. … Emotions play an important functional role, with the purpose of helping us to survive and adapt in complex and potentially hazardous physical and social domains (Panksepp & Biven, 2012). They aren't necessarily finely tuned but guide our behavior in directions evolution has taught us are wise."

What is proposed here is nowhere near a full model of emotion. It focuses on only the architectural aspects of how emotional states arise and affect cognition; and this it only does abstractly, not delving into the details of appraisal and dimensional models. It also has nothing to say at this point about such topics as how emotional states are reflected in external expressions. Still, the intent is to take a significant step in considering how emotion relates to architectures that align with the Common Model.

The next section provides background on the Common Model and how we arrived at this proposal. The subsequent two sections provide more details on two new modules that are proposed for inclusion into the Common Model – one for *emotion* and one for *metacognitive assessment* – and how they interact with the rest of the model. The final section summarizes what has been proposed here.

## Background

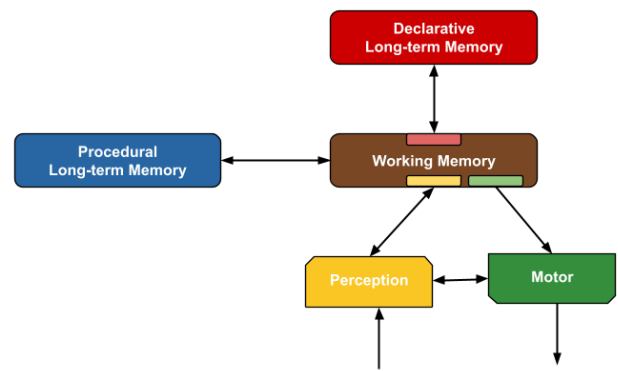

Figure 1: The Common Model of Cognition.

Figure 1 shows the basic structure of the Common Model. It comprises a central working memory, two long-term memories, and perception and motor modules. Working memory (WM) represents the current situation. Procedural long-term memory, which consists generically of rule-like structures, has direct access to all of WM. The other modules interact with it through dedicated buffers. Declarative long-term memory here does not (yet) distinguish between semantic and episodic knowledge. The perception and motor modules are minimally defined.

This figure is accompanied by sixteen assumptions about how it all works, divided up according to whether they bear on: (A) structure and processing; (B) memory and content; (C) learning; or (D) perception and motor control. Key assumptions, for example, include: (A3) there is significant parallelism both within and across the modules; (A4) sequential behavior arises from a cognitive cycle operating at ~50 msec in humans; (B1) long-term memories contain symbolic data with associated quantitative metadata; (B2) global communication occurs via WM; and (C2) learning occurs incrementally as a side effect of performance.

A broad survey of cognitive architectures can be found in Kotseruba and Tsotsos (2020), including examples of architectures with aspects of emotion. The proposal here, however, grew more directly out of an earlier analysis of the relationship of emotion to the Common Model (Larue et al., 2018) that later fed into a virtual workshop on the topic in June 2022. Out of the final session of that workshop came an initial consensus (Figure 2).

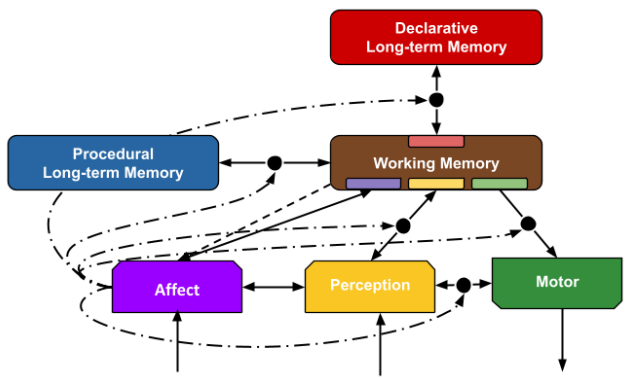

Figure 2: A version of the initial CMC+Emotion synthesis from the 2022 workshop.

The core of this figure is the pre-existing Common Model. Added to it is an *affect* module intended to capture key aspects of emotion. It receives input from physiology, to react to bodily state; from perception, to invoke immediate reactions to the state of the world without requiring direct cognitive participation; from its WM buffer, to invoke reactions to the state of the cognitive system; and from WM more broadly, and likely more diffusely, to set the context for affective processing. In return, the affect module provides input to its WM buffer, to enable reasoning by the cognitive system about its results; it filters communications between the other modules and WM; and it directly affects perception (see, e.g., Zadra & Clore, 2011).

In this model, reasoning about emotion – that is, the "cold" aspects of emotion, whether concerning oneself or others – is presumed to occur via standard cognitive processing within the original modules of the Common Model. The higher-level aspects of appraisal theory (e.g., Marsella & Gratch, 2009; Moors et al., 2013; Scherer, 2001) would thus fit here. The "hot" aspects of emotion are presumed to be the province of the physiological system that is not shown but that provides input to the affect module (e.g., Dancy, 2013). The model provides scope for both of these aspects of emotion but has nothing further to say about either of them.

What was termed the "warm" aspects of emotion at the workshop are the architectural facets of emotion processing, such as processing within the affect module itself, how metacognitive assessment lays the groundwork for it, and its impact on the rest of the Common Model. As with cold and hot emotion, the model does not delve into the internals of these two warm modules, but it does propose extending the Common Model's architecture to include them. Still, given that warm emotion is architectural, it would make sense for future work on the Common Model to accrete further details about how they operate.

One example architectural precedent for key aspects of this proposal can be found in the Sigma cognitive architecture (Rosenbloom, Gratch & Ustun, 2015). Attention there differentially abstracts messages throughout the cognitive system, affecting both communication across modules and within them, driven by a combination of two low-level, architecturally computed appraisals – *desirability* and *surprise* – which are themselves a function of what is perceived, what is learned, and what is in WM. The results of these appraisals then arrive back in WM. Other examples include West and Young's (2017) proposal for a similar extension to the Common Model that accesses WM like procedural long-term memory while providing subsymbolic evaluations back to both long-term memories, and Smith et al.'s (2021) argument for activation in ACT-R to include emotion via an additive scalar term.

Although there was a sense of consensus coming out of that workshop with respect to something like Figure 2, it was the result of only a few days' work by a subset of the community that did not actually come together until the final session. It thus did not seem right to consider it by itself as an official consensus. And, even if it were to become such, and thus part of the Common Model, it was quite minimal. So, the first four co-authors on this article set out to push the model further before going back to the community to see if a more thorough consensus was reachable. The proposal here is the product of these deliberations.

## Outline of the Proposal

Figure 3 outlines this new proposal. The changes in module locations from Figure 2 are purely cosmetic, to simplify and deconflict the resulting diagram. Each of the remaining changes reflects refinements that have been made to the initial CMC+Emotion model in Figure 2. This is a complex figure that will be broken down further in the next two sections.

One change that may look cosmetic is relabeling the affect module as *emotion*. Significant consideration went into the question of exactly what function this module – which has at various times been labeled affect, emotion, or physiology – should serve, as well as how it connects to the other modules. It being labeled emotion in Figure 3 implies this concluded with the idea that while it senses physiology this module's role is to generate emotional vectors, such as <valence, arousal> or more extended vectors, that are central to dimensional models of emotion (e.g., Juvina, Larue & Hough, 2018; Mehrabian & Russell, 1974; Rubin & Talarico, 2009), but without a commitment to the size or contents of such a vector.

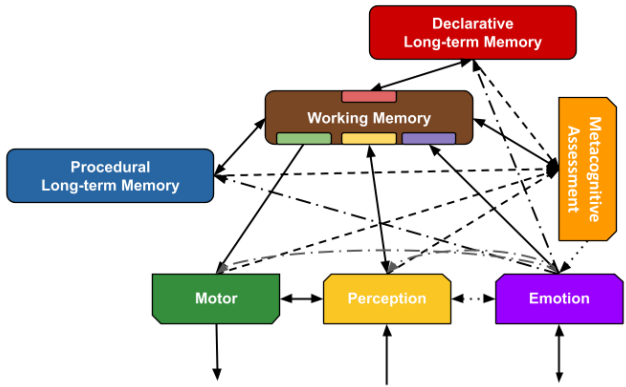

Figure 3: Refined CMC+Emotion synthesis that is the basis of the current proposal.

One major change was due to noticing that an important pathway, and associated module, was missing from Figure 2 that generates low-level appraisals based on observing the existing modules and the traffic among them, and transmits the results to the emotion module. This has taken the form here of a metacognitive assessment module that is discussed after the emotion module.

A second major change is that the emotion module here not only filters communications between other modules and WM but also affects how the modules themselves operate. One simple example of the latter is how emotions may yield rewards for reinforcement learning in procedural long-term memory (e.g., Marinier & Laird, 2008), or somatic markers in declarative long-term memory (Damasio, 1994).

More broadly, extensive evidence indicates that emotions can exert specific influences on memory storage and retrieval, affecting the processing of learned information. These include much-studied everyday effects such as state-dependent memory (e.g., Eich, 1995) and clinical effects such as post-event emotional and traumatic responses (Brewin, 2011) that are directly mediated via cortico-amygdala loops (e.g., Fadok et al., 2018; Grundemann et al., 2019). Several specific hypotheses about the nature of the effects of emotional content on memory storage, retrieval, and inference can be posed both behaviorally and neurally to clarify features of emotion-memory interactions (e.g., Fadok et al., 2018; Janak & Tye, 2015; Kesler, 2001; Saarimaki, 2016).

The emotion and metacognitive assessment modules can both be seen as analogous to the perception module, although they perceive physiology and the cognitive system respectively rather than the external environment. Similar analogies are also conceivable between these new modules and the motor module when their ability to act on their environments is considered.

The beginnings of an attempt has been made to determine if a consensus was reachable around this new proposal, involving two separate emails to the workshop attendees requesting input from them on it, one informal, as free text, and one structured more formally as a questionnaire, but this process proved insufficient to yield a consensus even among the workshop attendees, so no attempt has yet been made at achieving a broader community consensus.

This material is therefore presented as a proposal for further consideration rather than as an agreed-upon extension to the Common Model. The other two co-authors on this paper were workshop attendees who agreed to join in this latest stage of proposal refinement and presentation.

## Emotion Module

Figure 4 is a simplified version of Figure 3 that eliminates metacognitive assessment. This version is much like Figure 2, but for two key extensions. First, there is a connection from the emotion module back down to physiology for emotion vectors to affect physiology; for example, when a cognitively identified threat – such as a verbal threat – requires the body to prepare to respond. Second, the dot-dash arrows from the emotion box now point to the junctions between the non-WM Common Model modules and their links to WM. This is to indicate that not only can the vectors from the emotion module filter communication along these links, but they can also modify how these modules work. Although the model does not specify how this happens, examples may include altering how procedural memory selects actions to execute and how declarative memory determines what knowledge to retrieve.

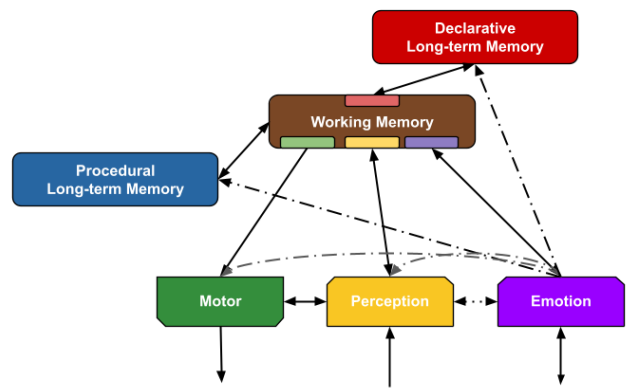

Figure 4: A version of Figure 3 simplified to omit metacognitive assessment.

In Figure 4, as in Figure 2, there is a WM buffer associated with the emotion module. The arrow to this buffer enables the vectors generated by the module to appear in working memory. In Common Model fashion, these vectors could arrive as symbolic data identifying the dimensions of the vector with quantitative metadata that specifies their values.

The return arrow provides cognitive input to the emotion module, including high-level appraisals. As discussed further in the next section, low-level appraisals follow a different route, via the metacognitive assessment module. With input to the emotion module that includes both high-level and low-level appraisals, and output from it in the form of vectors, this proposal implies a connection between appraisal and vector models in which the former are inputs to creation of the latter. However, vector generation also depends on inputs from physiology and perception, and possibly other content in WM.

Two simple examples of how this might proceed based just on appraisals, albeit at opposite ends of the vector-length spectrum, are (1) generation of large vectors by assigning one slot to each appraisal; or (2) generation of intensity and valence pairs by aggregating over the values of all appraisals for the former and differentially over positive and negative appraisals for the latter. Under the second option, physiological inputs might combine straightforwardly with the two values derived from appraisals.

Figure 4 includes a bidirectional arrow between emotion and perception, although it is shown as a dotted line because it remains unclear whether a direct return path from emotion to perception is needed in addition to the curved arrow that already indicates emotional modulation of perception.

## Metacognitive Assessment Module

In attempting to include appraisals and their relationship to emotions, we ended up modeling them generically as aspects of metacognition – where the cognitive system operates on itself – a large-scale topic of its own on which an overall consensus has not yet been reached with respect to the Common Model, although Kralik et al. (2018) did begin exploring this question.

The particular point of interest here is that some forms of low-level appraisals, such as surprise and familiarity, can be thought of as metacognitive assessment that is grounded in fixed, architectural sensors that observe what is happening within the overall cognitive system. As one simple example, both surprise and familiarity are computed architecturally in Sigma based on monitoring its learning process.

This approach would put such appraisals in the same category as, for example, a sensor for *feeling of knowing* that assesses when declarative memory will be able to retrieve an appropriate memory given a cue (Nhouyvanisvong & Reder, 1998). It would also put them in the class of "warm" aspects of emotion.

Figure 5 shows a version of Figure 3 that includes the metacognitive assessment module for low-level appraisals, but which is simplified via the removal of the dot-dash arrows from the emotion module to the relevant Common Model modules.

In this figure it can be seen how the low-level appraisals from this new module act as inputs to the emotion module. However, it remains unclear in general whether these appraisals should arrive directly from the metacognitive assessment module via the dotted arrow between the two modules, or whether this path can be omitted given the existence of the path via solid arrows that traverses WM.

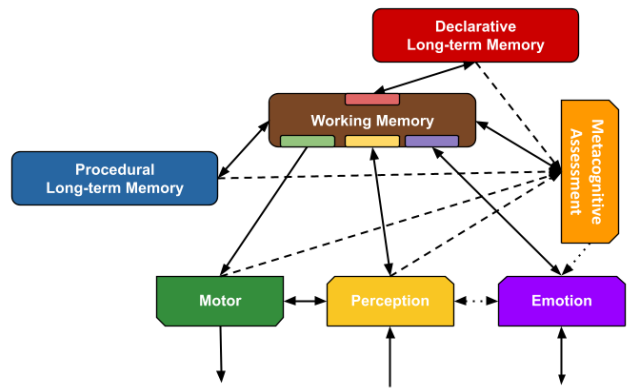

Figure 5: A version of Figure 3 simplified to omit the dot-dash arrows from the emotion module to the non-WM Common Model modules.

High-level "cold" appraisals, such as causal attribution, are considered to be essentially cognitive in nature, although Figure 5 abstracts over whether this form of metacognition occurs within (possibly a recursion on) the same cognitive system (e.g., Rosenbloom, Laird & Newell, 1988) or via a distinct metacognitive system (e.g., Cox, Oates. & Perlis, 2011; Sun, Zhang & Matthews, 2006). It makes sense to defer attempting to resolve such a question until a full exploration is begun of how to extend the Common Model to metacognition.

As shown in Figure 5, the metacognitive assessment module receives input from each of the other non-WM modules in the original Common Model. That these connections are from junctures between these other modules and their links to WM is intended to indicate that the metacognitive assessment module can sense both their communication with WM and what is going on within them, although the figure abstracts over exactly what is sensed. None of this sensing of the cognitive system is reflected in Figure 2, but it is intended to effectively be the inverse of how the emotion module acts upon these junctures.

There are no arrows back from the metacognitive assessment module to the other modules, which might be expected in a full analysis of metacognition, but these do not appear necessary for emotional metacognitive assessment. Instead, what feedback does occur goes through the emotion module before it reaches them. Whether direct backward connections are ultimately needed in addition to this route through the emotion module remains to be seen.

Although metacognitive assessment is shown in Figure 5 as a separate module, it is not yet clear whether it should truly be considered a module on its own versus there being merely bits of it distributed across the other modules and connections, where it is presumed that the sensing actually occurs. It is shown as a module here to leave open the possibility of architectural across-module appraisals – such as the earlier Sigma example in which attention is based on both surprise and desirability – rather than assuming that all low-level assessments are specific to one module or that all combinations of them happen cognitively. However, the necessity of such a possibility remains as another open question.

An arrow is shown in Figure 5 from metacognitive assessment to WM to make low-level appraisals accessible to cognition in support of higher-level appraisal, including complex across-module appraisals, as well as other relevant cognitive processing. The reverse arrow, from WM back to the module indicates WM affecting metacognitive assessment.

It is left open whether these interactions are as unconstrained as those between procedural long-term memory and WM or whether they are constrained to go through a module-specific buffer, as is the case with the other modules. However, if the arrow from WM to the metacognitive assessment module is unconstrained, it may be able to substitute for Figure 2's arrow from WM to the affect module, supporting a flow from all of WM, through metacognitive appraisal, to emotion.

The previous section raised the question of whether a direct connection is required from emotion to perception. In the current context, the existence of a pathway from perception, through metacognition, to emotion raises the reverse question, as to whether a direct link from perception to emotion is necessary when this slightly less direct path already exists.

## Summary

The proposal presented here for extending the Common Model of Cognition to aspects of emotion includes a new emotion module that can affect the workings of the existing non-WM modules as well as filter their communications with WM. It also includes a new metacognitive assessment module that can perceive WM plus the workings of the existing non-WM modules and their communications with WM. These two modules, plus links between them and between the emotion module and physiology, comprise the core of the proposed model, as outlined in Figure 3.

This model is of course incomplete in many ways with respect to the full complexity of emotion. It shares the Common Model's natural abstraction and incompleteness in terms of only including aspects about which there is a consensus, although here this is in terms of what might become a consensus. Thus, this article still reflects only a beginning of a beginning at extending the Common Model to emotion, even while building on multiple earlier efforts in this direction. Still, given the importance of the connection between cognition and emotion, it hopefully provides a basis for a wider discussion of what should be added to the Common Model in support of extending it to emotion.

## Acknowledgments

We would like to thank everyone who participated in the June 2022 workshop and who thus laid the groundwork for the initial model shown in Figure 2.